\documentclass[11pt]{article}
\usepackage{caption}
\usepackage{subcaption}
\usepackage{amsfonts}
\usepackage{bbm}
\usepackage{amsmath}
\usepackage[ruled,vlined,linesnumbered]{algorithm2e}
\usepackage[section]{placeins}
\usepackage{epsfig}
\usepackage{amssymb}
\usepackage[numbers]{natbib}
\usepackage{graphicx}
\usepackage{amsthm}

\usepackage{booktabs}
\usepackage{etoolbox}
\usepackage[round-mode=places,detect-weight=true,detect-inline-weight=math]{siunitx}

\theoremstyle{definition}

\SetKwInput{KwInput}{Input}                
\SetKwInput{KwOutput}{Output}              

\begin{document}




\title{A Non-Linear Neuron Based Detection of Isolated Pixels in Binary and Grayscale Images using Contrast Sensitive Receptive Fields}
\author{Nassir Mohammad\thanks{Email: nassir.mohammad@airbus.com} \thanks{Project code is available at https://github.com/M-Nassir/isolated\_pixel\_detection} \\ 
\emph{Cyber Innovation Lab, VCX, Airbus, Newport, UK}}
\maketitle
\begin{abstract}
Identifying isolated points is important in image processing applications such as medical imaging, astronomy and quality control management. Other domains, such as cybersecurity, also present challenges that can be framed as image processing problems. One example of particular interest is the identification of anomalous single nodes in spatially organised networks where groups of nodes in different regions share similar feature values. This task can involve both binary and more complex grayscale images. However, existing methods face limitations: template matching is infeasible for grayscale images, while 2nd order derivative based methods are highly sensitive to noise and require user-specified thresholds. To overcome these issues, a novel method is proposed for detecting meaningful single-pixel deviations in images. This approach modifies and extends a neuron model, originally designed for anomaly detection, to operate on spatially diameter limited receptive fields that incorporate excitatory and inhibitory regions. The result is a method that is free from user-specified thresholds and parameters, and can be applied to both binary and grayscale images, providing an effective, robust and efficient solution.
\end{abstract}


%
\FloatBarrier
\section{Introduction} 
\label{section:introduction} 

The identification of low-level meaningful features in images is a crucial part of many computer vision applications. These features are typically categorised as isolated points, lines, bars, and edges and play a vital role in solving image processing problems. Despite its importance, feature detection remains a difficult problem due to the numerous challenges that can and $do$ occur in images, including noise, occlusions, and varying lighting conditions--a reality that has been acknowledged since the earliest days of computer vision \cite{Mar82}.

The present work focuses on the simplest of image processing problems; the detection of single pixels that exhibit significant deviations from their local surroundings. Some examples where this has application include the detection of porous points in aircraft components, the appearance of tiny defects in controlled industrial imagery, and scanning or noise artifacts in medical and document imaging. Moreover, by transforming a problem into that of an image processing task, this can be applied to the detection of malicious or rogue nodes in spatially organised networks where nearby nodes share similar characteristics.

The detection of abrupt intensity changes at single pixel locations surrounded by homogenous areas can be carried out in a number of ways such as exact template matching (for binary images) and using second order derivatives (for both binary and grayscale images) where sharp discontinuities cause high levels of filter responses. These approaches have been used since the earliest days of image feature detection and still continue to be the basis of the most widely used detectors \cite{gonzalez2008digital}. However, issues with both the approaches appear when dealing with grayscale images. Template matching becomes infeasible due to the sheer number of possibilities and decisions on which to characterise templates. On the other hand, derivative based approaches are sensitive to imperceptible noise, require thresholding parameters or tuning, and do not always perform adequately in real-world applications. 

The main contribution of this work is the introduction of a novel approach for the detection of image percepts using non-linear neurons performing unsupervised anomaly detection within spatially local receptive fields, together with elements of retinal lateral inhibition. Such an approach no longer uses derivatives as a measure of change, but the unexpected nature of responses within a receptive field, and thus does away with the problem of thresholding by adapting to the input data. The performance of this approach hinges on the acceptance of the definition of an anomaly proposed by Mohammad \cite{nassir2021anomaly}, and the use of the \emph{perception} algorithm that detects anomalies without specification of user-parameters. Thus, with an initial realisation of the neuron model operating over spatial inputs, an effective unsupervised parameter-free method is provided for the detection of isolated single pixels in binary and grayscale images--tackling one of the simplest of image processing tasks. 

The rest of the paper is laid out as follows: Section \ref{section:prior_art} describes the prior art in detecting isolated pixels. Section \ref{section:neuron_based_solution} describes the non-linear neuron based approach and model for detecting isolated pixels. Section \ref{section:example_results} gives some examples and compares the method proposed in this paper with that of template matching and second order derivatives. Section \ref{section:conclusion_and_future_work} concludes the paper and outlines future work.

\section{Prior Art}
\label{section:prior_art}
This section describes two of the main approaches used to detect isolated pixels in binary and grayscale images.

\subsection{Template Matching: the Hit or Miss Operator}
\begin{figure}
\centering
\begin{subfigure}[b]{0.4 \textwidth}
  \includegraphics[width=\textwidth]{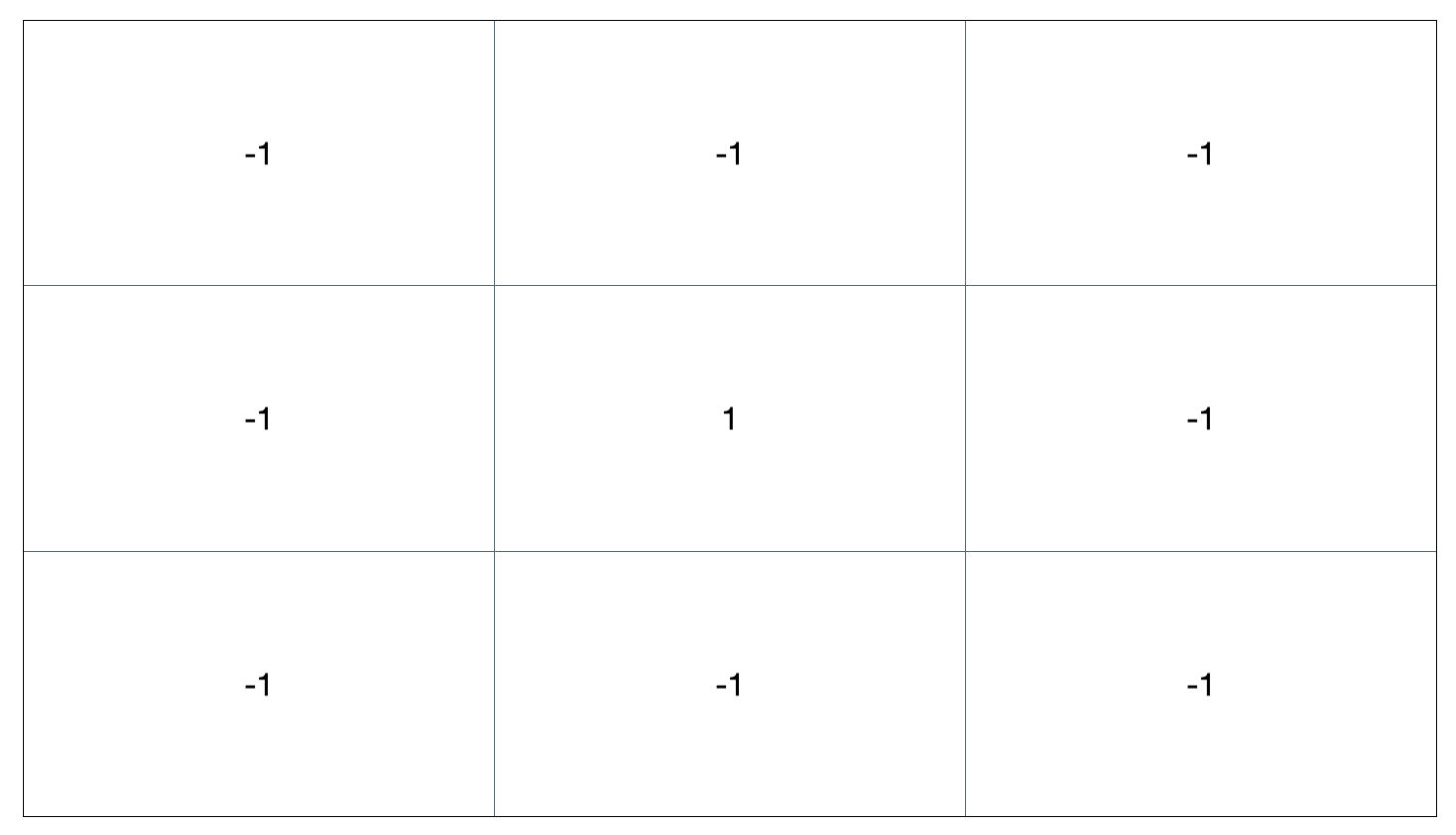}
  \caption{}
  \label{fig:isolated_pixel_mask_a}
\end{subfigure}
\begin{subfigure}[b]{0.4 \textwidth}
  \includegraphics[width=\textwidth]{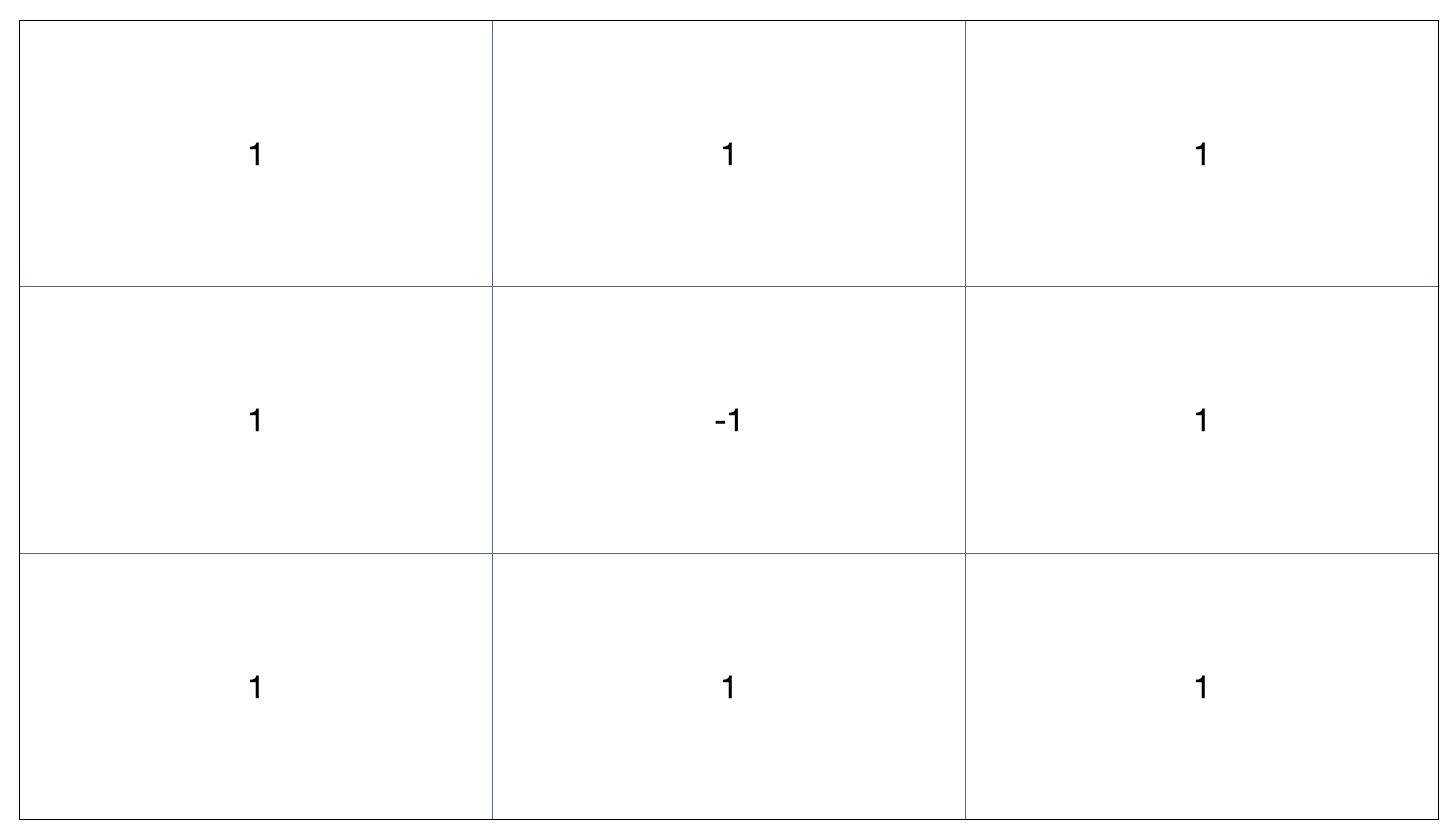}
  \caption{}
  \label{fig:isolated_pixel_mask_b}
\end{subfigure}
\caption{Hit-or-miss masks for detecting isolated pixels. (a) Mask designed to detect isolated foreground pixels surrounded by background pixels. (b) Mask for detecting isolated background pixels surrounded by foreground pixels.}
\label{fig:isolated_pixel_mask}
\end{figure}

The detection of isolated pixels in binary images can be effectively performed using the hit-or-miss transform, a binary morphological operation designed to detect specific patterns of pixels \cite{gonzalez2008digital}. This method involves applying a structuring element to the binary image, as shown in Figure \ref{fig:isolated_pixel_mask}(\subref{fig:isolated_pixel_mask_a})  for the detection of isolated white pixels, and Figure \ref{fig:isolated_pixel_mask}(\subref{fig:isolated_pixel_mask_b})  for the detection of isolated black pixels. In this process, the structuring element is slid across the image, and at each position, the mask is checked for an exact match with the underlying pixel values. If the mask exactly matches the pixel values under it, the pixel in the output image corresponding to the central pixel of the structuring element is set to the value defined by the structuring element (e.g., the foreground color). If there is no match, the pixel is set to the background colour. This results in a filter response image that highlights only the detected isolated pixels.

The hit-or-miss transform is limited to binary images where it is naturally constrained by there being only two possible pixel values and isolated pixels corresponding to two fixed patterns. In grayscale images, where pixels can have a wide range of values, there are many possible patterns to consider. For example, if we consider a small 8-bit $3\times3$ patch of pixels, each taking $256$ possible values, then there are $256^9$ possible patterns. This results in an impractically large number of templates to apply, making the template matching approach infeasible for detecting isolated pixels. 

\subsection{Second Order Derivatives: the Laplacian Filter}
\begin{figure}
\centering
  \includegraphics[width=0.4 \textwidth]{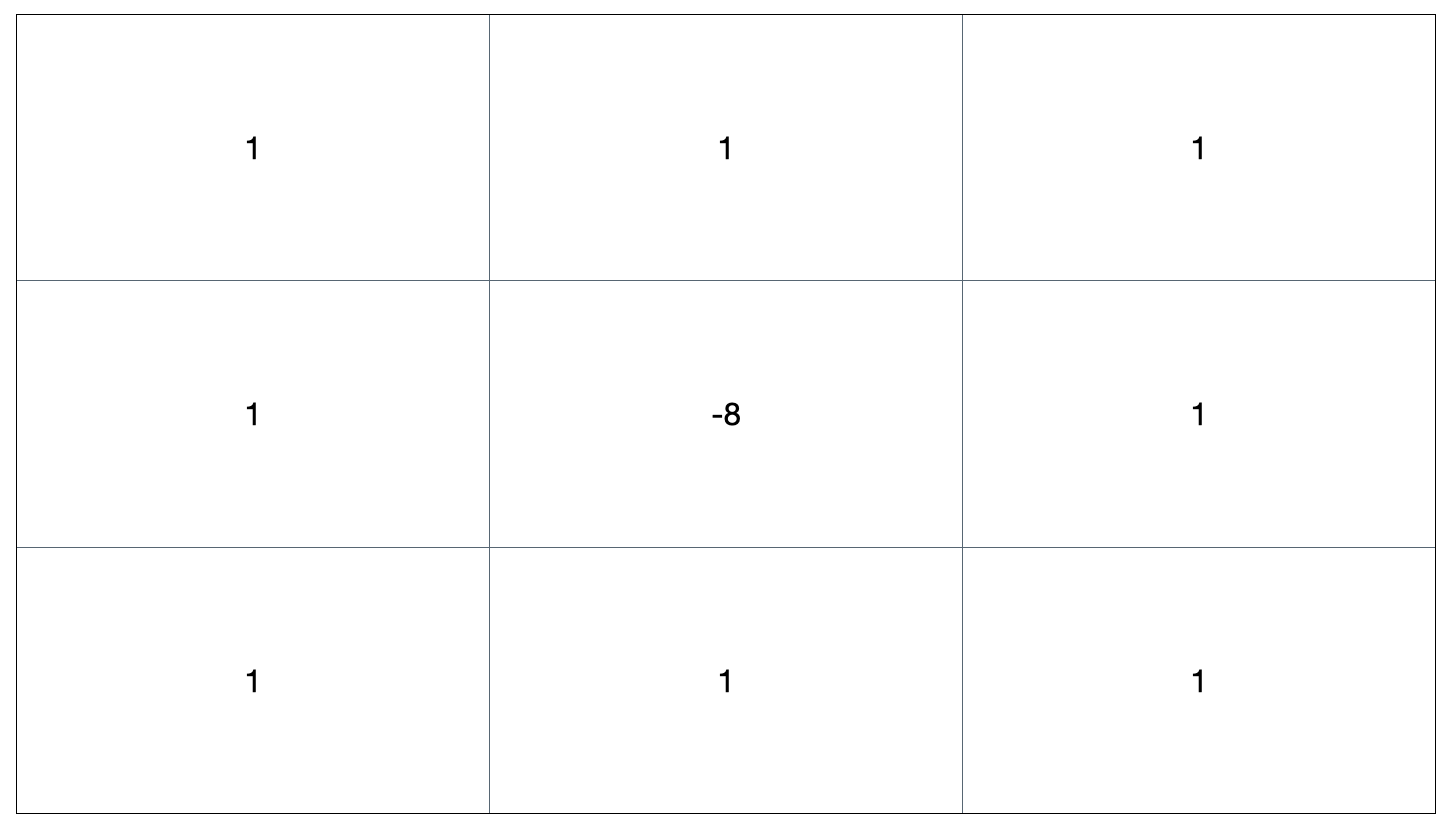}
  \caption{Laplacian mask used for detecting isolated points in an image. This mask highlights regions of rapid intensity change by calculating the second-order derivatives, making it sensitive to both noise and fine details in the image.}
\label{fig:laplacian_mask}
\end{figure}

Abrupt local changes in image grayscale intensity values can be detected using first and second-order derivatives, with the latter being particularly effective for detecting isolated pixels \cite{gonzalez2008digital}. Derivatives are approximated using differences between pixel values. Specifically, any approximation of the second derivative must be zero in areas of constant intensity, zero at the onset and end of an intensity step or ramp, and zero along intensity ramps. 
The Laplacian,
\begin{equation}
 \nabla^2 f(x,y) = \frac{\partial^2 f}{\partial x^2} + \frac{\partial^2 f}{\partial y^2}
\end{equation}
is particularly well-suited to detecting isolated pixels. The partial derivatives required for computing the Laplacian are approximated using the filter shown by Figure \ref{fig:laplacian_mask}.
The intuition for using this filter is that isolated pixels exhibit intensity values that significantly differ from their surrounding pixels, making them stand out after processing. Note that in areas of constant intensity the filter will respond with a zero output since the coefficients of the mask sum to zero. To identify the isolated pixels, the absolute value of the filter response must be taken, and if the resulting value is greater than a threshold $T$, determined by domain or empirical knowledge, the point is considered an isolated pixel. Running the filter across the image yields a binary filter response image:
\begin{equation}
  g(x,y) =
    \begin{cases}
      1 & \text{if $|R(x,y)|$ $\geq$ T}\\
      0 & \text{otherwise}
    \end{cases}       
\end{equation}
where $R(x,y)$ is the response of the filter centered on pixel $(x,y)$.

The use of second-order derivatives for detecting isolated pixels presents several challenges. One key issue is the need to define an appropriate threshold to determine whether a point is truly isolated. An improperly set threshold can result in numerous false positives or false negatives. While controlled environments may allow for precise parameter tuning, many applications often involve varying intensity changes, making it difficult to specify a single threshold that works consistently across different images or even within a single image. Another significant drawback is the high sensitivity of derivatives to noise, where even noise imperceptible to the human eye can significantly affect the derivative output, while the second-order derivative is even more sensitive \cite{gonzalez2008digital}. This leads to difficulties in the detection of isolated points in the presence of noise. 


\section{The Non-linear Neuron Detector}
\label{section:neuron_based_solution}
\begin{figure}
  \centering 
  \includegraphics[width=0.8\textwidth]{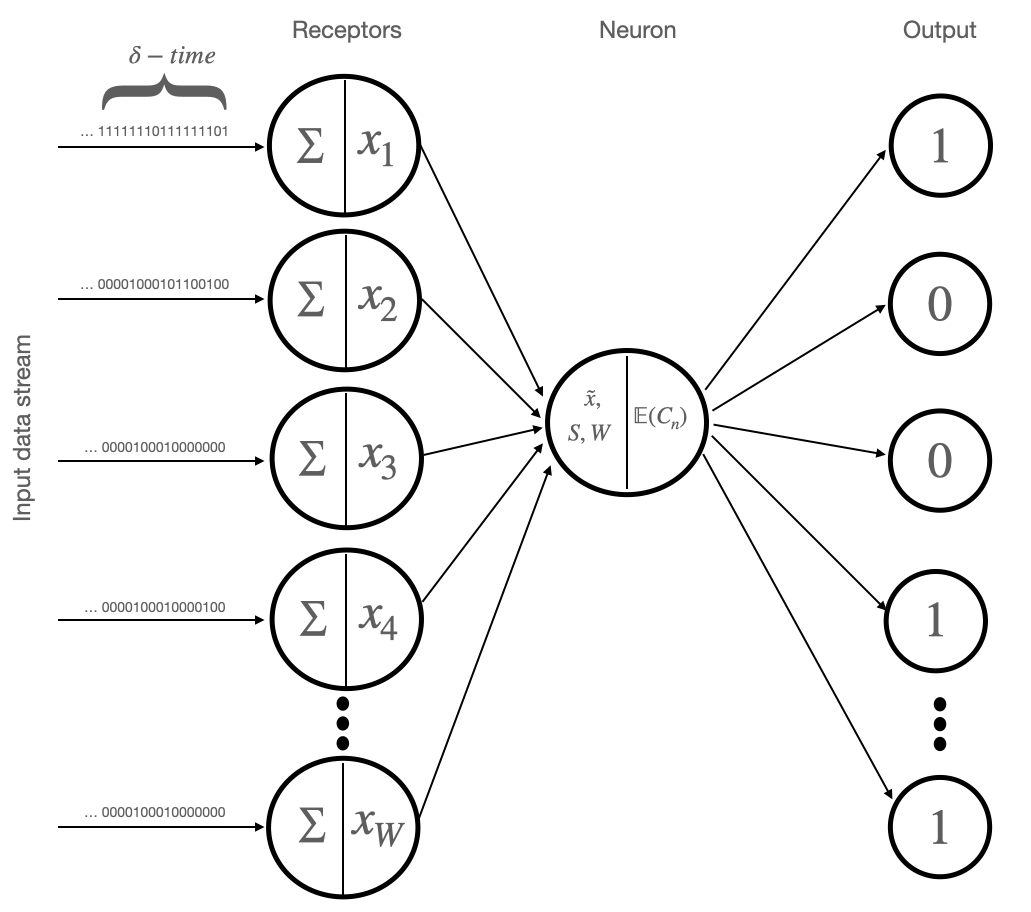} 
  \caption{The non-linear neuron model proposed by \citet{mohammad2022net} for the detection of anomalies.} 
  \label{fig:single_neuron_model} 
\end{figure}

A non-linear neuron model introduced by \citet{mohammad2022net} (illustrated in Figure \ref{fig:single_neuron_model}) is designed to detect anomalies assumed to have been generated by data streams of binary $indicators$ arriving at each receptor node. The core computation within the neuron node employs the perception algorithm developed by \citet{nassir2021anomaly}. This model is particularly effective for unsupervised anomaly detection because it adapts automatically to the underlying input distribution, removing the need for manual parameter tuning. In addition, it performs robustly on both small and large datasets, reinforcing the soundness and generality of its underlying design philosophy.

\begin{figure}
  \centering 
  \includegraphics[width=0.8\textwidth]{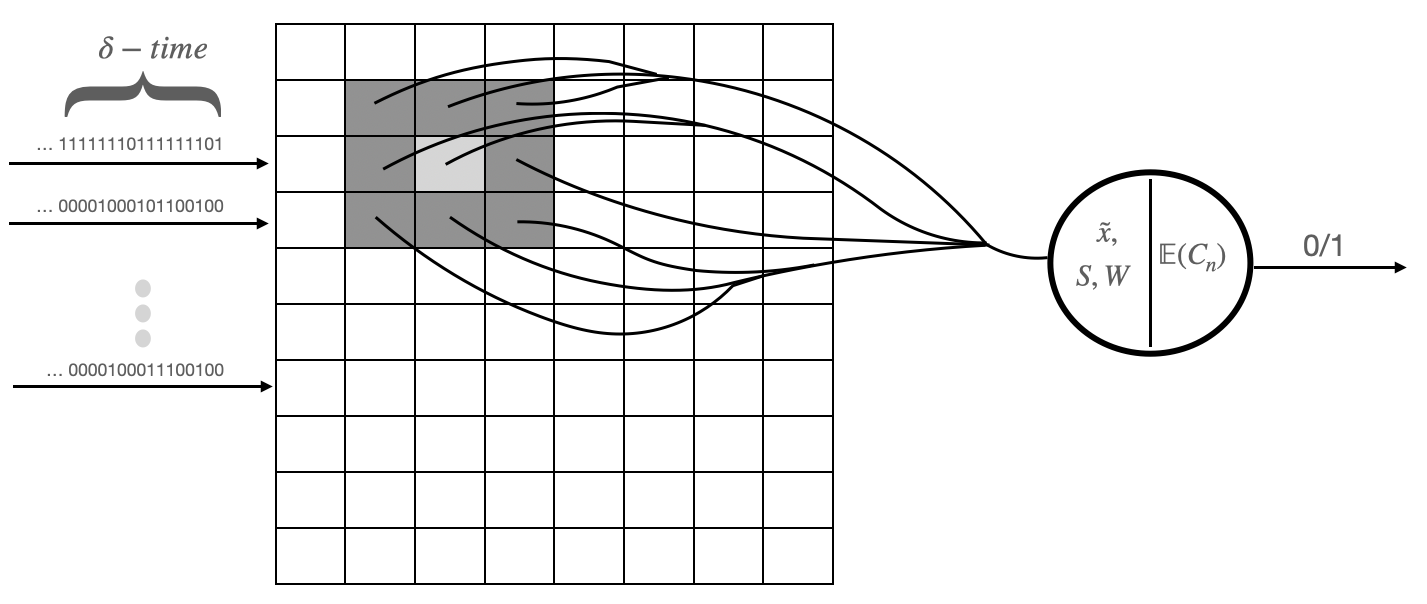} 
  \caption{The non-linear neuron model designed to operate on spatially diameter limited receptive fields over an entire image. The $3\times3$ pixel receptive field features a central excitatory region (lighter-shaded pixel) surrounded by an inhibitory region (darker-shaded pixels). While the neuron can detect anomalies at any position, it only fires when an anomaly is present exclusively in the excitatory region and none are detected in the inhibitory region.}
  \label{fig:neuron_model_parallel} 
\end{figure}

Inspired by receptive fields and retinal ganglion cell processing in the human retina, the present work modifies and extends the neuron model. In the fovea, individual ganglion cells communicate with as few as five to ten photoreceptors or up to thousands in the peripheral, some sensitive to even single photons, enabling them to detect minute light changes. The receptive fields are highly organised and localised, typically consisting of excitatory and inhibitory regions enabling the bilogical neuron to effectively discern contrast changes. Drawing on these biological properties, the neuron model is modified to operate on spatially diameter limited receptive fields (presently on $3\times3$ pixel regions), where it operates over the input values using the perception algorithm. The neural computation, depicted in Figure \ref{fig:neuron_model_parallel}, involves calculating the median $\tilde{x}$ of all its input intensity values $x_i$ (integrated over a stream of indicators for interval $\delta$ which can simply be the largest intensity value), a sum $S$ of the magnitude differences between the $x_i$ and the median $\tilde{x}$, and the number of input connections $W$ (with $W = 9$ for a $3 \times 3$ window). For each $n = |x_i - {\tilde{x}}|$ the following expectation is computed, where any $x_i$ satisfying the equation is considered an anomaly by definition since its occurrence is unexpected to occur even once, but it has occurred:
\begin{equation}
  \mathbb E(C_{n}) = \binom{S}{n} \frac{1}{W^{n-1}} < 1
\end{equation} 
However, the central position of the receptive field is designated as an excitatory region, while the surrounding area is considered inhibitory. Taking the stance that inhibition is designed to produce specialisation, a simple mechanism is employed: the neuron fires only when the excitatory region is active (i.e., an anomaly detected), and any activity (anomalies) within the inhibitory region suppresses the neuron's response entirely. As a result, the neuron must $only$ detect an anomaly in the central position for it to fire. This mechanism ensures selective response to anomalies localised in the excitatory zone.

To detect isolated pixels in an image, the neuron model is applied across all possible $3\times3$ windows in parallel, with the response captured to create a filter response map (e.g., Figure \ref{fig:binary_images_with_filter_responses}(b)). This method circumvents challenges associated with implementing exact template matching, calculating derivatives, and the need for threshold selection.

\section{Example Results}
\label{section:example_results}
This section gives examples of detecting isolated pixels in binary, artificial and natural grayscale images using template matching (the hit-or-miss transform), second order derivatives (the Laplacian), and the non-linear neuron model proposed in this paper. (Images are best viewed close up in their digital format due to the small nature of isolated pixels.)

\begin{figure}
  \centering
  \begin{subfigure}{0.3\textwidth}
    \centering
    \includegraphics[width=\linewidth]{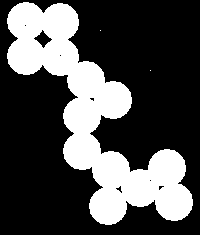}
    \caption{}
    \label{fig:circles_matlab}
  \end{subfigure}
  \hspace{0.5cm} 
  \begin{subfigure}{0.3\textwidth}
    \centering
    \includegraphics[width=\linewidth]{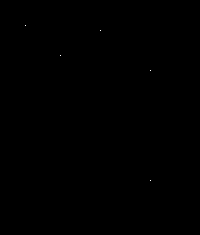}
    \caption{}
    \label{fig:filter_response_circles}
  \end{subfigure}
  
  \vspace{0.5cm} 
  
  \begin{subfigure}{0.35\textwidth}
    \centering
    \includegraphics[width=\linewidth]{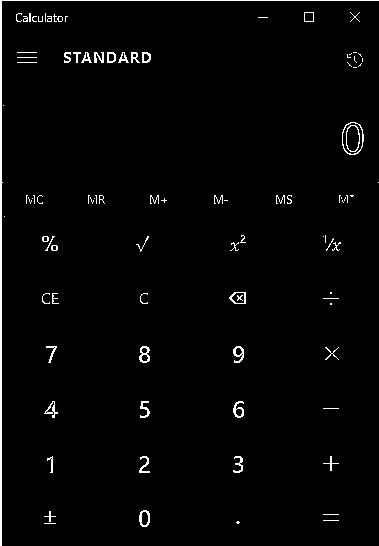}
    \caption{}
    \label{fig:calc}
  \end{subfigure}
  \hspace{1cm} 
  \begin{subfigure}{0.35\textwidth}
    \centering
    \includegraphics[width=\linewidth]{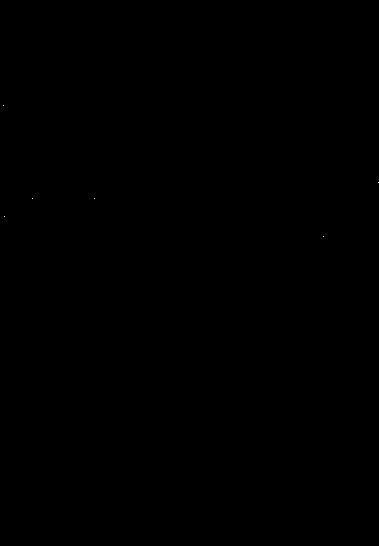}
    \caption{}
    \label{fig:filter_response_calc}
  \end{subfigure}
  
  \vspace{0.5cm} 
  
  \begin{subfigure}{0.3\textwidth}
    \centering
    \includegraphics[width=\linewidth]{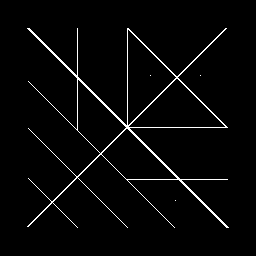}
    \caption{}
    \label{fig:crosses}
  \end{subfigure}
  \hspace{0.5cm} 
  \begin{subfigure}{0.3\textwidth}
    \centering
    \includegraphics[width=\linewidth]{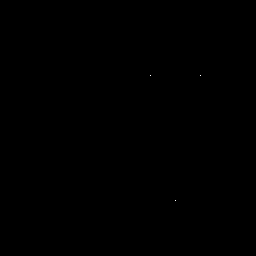}
    \caption{}
    \label{fig:filter_response_crosses}
  \end{subfigure}
  
  \caption{Original binary images with isolated pixels (left column) and their corresponding filter response maps (right column). All the methods detected the isolated pixels inside the border of the images correctly.}
  \label{fig:binary_images_with_filter_responses}
\end{figure}

%
%
Figure \ref{fig:binary_images_with_filter_responses}(a), (c) and (e) display binary images containing isolated pixels, while Figure \ref{fig:binary_images_with_filter_responses}(b), (d) and (f) display the filter responses after applying the hit-or-miss transform, Laplacian followed by thresholding ($T=0.9$), and the neuron model, respectively. All methods successfully detect the isolated pixels within the image borders. Both template matching and the Laplacian-based approach perform well for this task—the binary nature of the images constrains the problem, making template matching feasible, and the Laplacian method effective with a threshold of $0.9$.

\begin{figure}
  \centering
  \begin{subfigure}{0.3\textwidth}
    \centering
    \includegraphics[width=\linewidth]{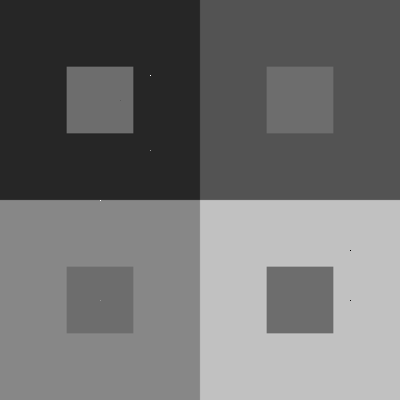}
    \caption{}
    \label{fig:circles_matlab}
  \end{subfigure}%
  \hfill
  \begin{subfigure}{0.3\textwidth}
    \centering
    \includegraphics[width=\linewidth]{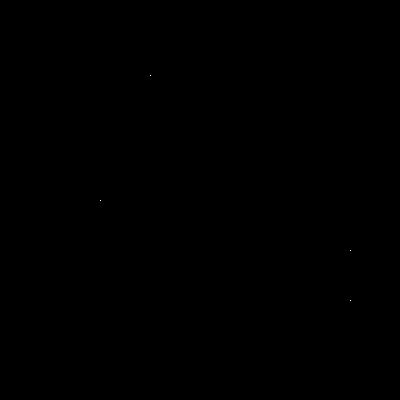}
    \caption{}
    \label{fig:filter_response_circles}
  \end{subfigure}
  \hfill
  \begin{subfigure}{0.3\textwidth}
    \centering
    \includegraphics[width=\linewidth]{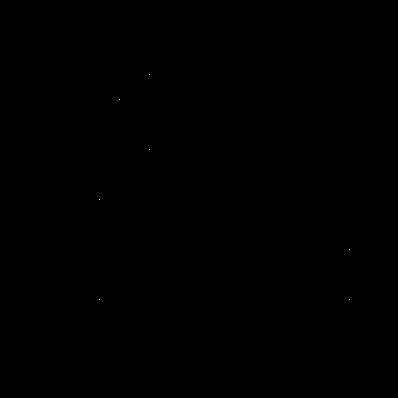}
    \caption{}
    \label{fig:calc}
  \end{subfigure}%
  
  \vspace{0.5cm}
  \begin{subfigure}{0.3\textwidth}
    \centering
    \includegraphics[width=\linewidth]{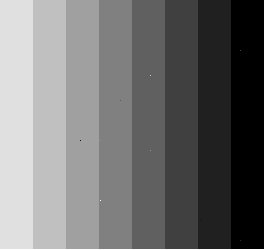}
    \caption{}
    \label{fig:calc}
  \end{subfigure}%
  \hfill
  \begin{subfigure}{0.3\textwidth}
    \centering
    \includegraphics[width=\linewidth]{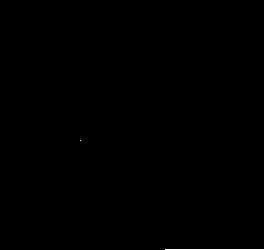}
    \caption{}
    \label{fig:filter_response_calc}
  \end{subfigure} 
  \hfill
  \begin{subfigure}{0.3\textwidth}
    \centering
    \includegraphics[width=\linewidth]{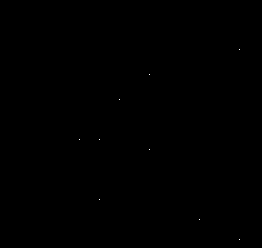}
    \caption{}
    \label{fig:filter_response_calc}
  \end{subfigure} 
  \caption{Comparison of original grayscale images with 7 isolated pixels added to the image of sqaures and 9 in the mach bands image (left column), Laplacian filter responses with $T=0.7$ (centre column), and neuron filter response maps (right column). The Laplacian method did not detect all isolated pixels and exhibited an unintended response along the bottom-right edge of the mach bands image. Adjusting $T$ did not necessarily improve the results. In contrast, the neuron model successfully detected all isolated pixels in both images parameter-free.}
  \label{fig:piecewise_grayscale_images_with_filter_responses}
  \end{figure}

%
%
The next level of complexity involves grayscale images featuring regions of constant intensity interrupted by lines or edges, as illustrated in Figure \ref{fig:piecewise_grayscale_images_with_filter_responses}, where isolated pixels of varying intensities have also been added to the images. The standard hit-or-miss transform becomes impractical here due to the exhaustive requirement of checking all possible intensity combinations. The Laplacian method necessitates a specified threshold which can require manual inspection of the images or knowledge of the isolated pixels. This reliance on known thresholds limits its practical application, particularly when these values are not consistently established. Even with a threshold set at $0.7$ of the maximum value, the Laplacian fails to detect all isolated pixels, and further adjustments to the threshold did not yield perfect results. In contrast, the neuron model operates effectively on such images without needing parameters, successfully distinguishing all isolated pixels from their background.

\begin{figure}[t]
  \centering
  \begin{subfigure}{0.3\textwidth}
    \includegraphics[width=\linewidth]{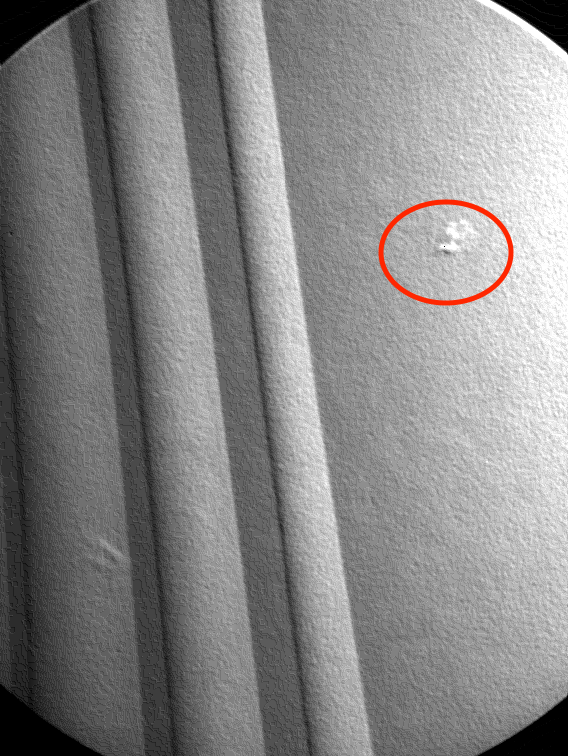}
    \caption{}
  \end{subfigure}%
  \hfill
  \begin{subfigure}{0.3\textwidth}
    \includegraphics[width=\linewidth]{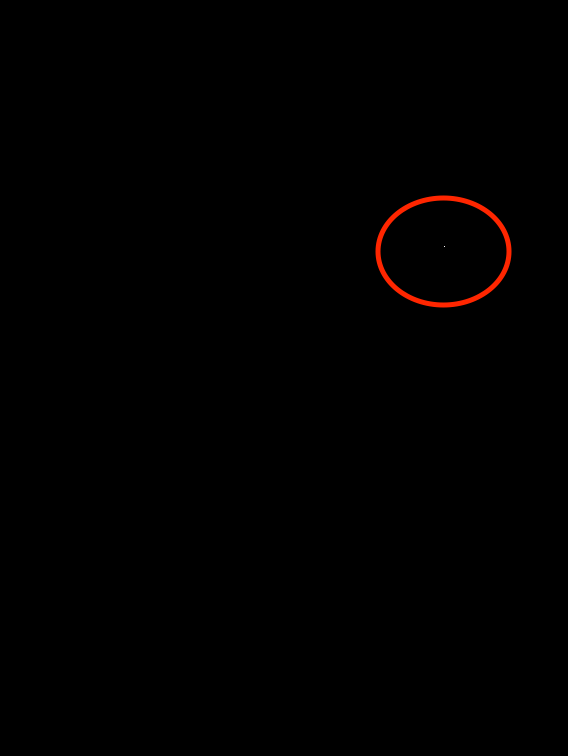}
    \caption{}
  \end{subfigure}%
  \hfill
  \begin{subfigure}{0.3\textwidth}
    \includegraphics[width=\linewidth]{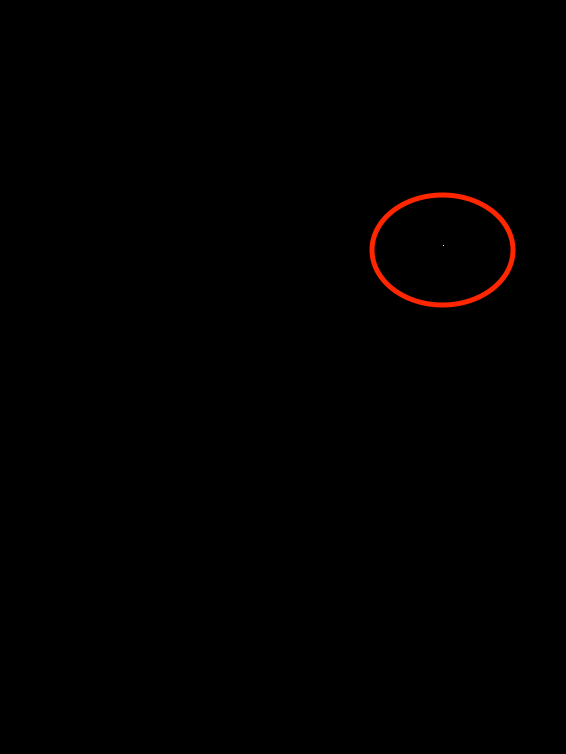}
    \caption{}
  \end{subfigure}

  \begin{subfigure}{0.3\textwidth}
    \includegraphics[width=\linewidth]{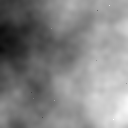}
    \caption{}
  \end{subfigure}%
  \hfill
  \begin{subfigure}{0.3\textwidth}
    \includegraphics[width=\linewidth]{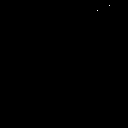}
    \caption{}
  \end{subfigure}%
  \hfill
  \begin{subfigure}{0.3\textwidth}
    \includegraphics[width=\linewidth]{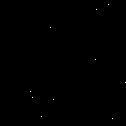}
    \caption{}
  \end{subfigure}
  \caption{Top row: (a) X-ray of a turbine blade with a tiny defect, (b) Laplacian response at $T=0.9$, and (c) neuron-model response after Gaussian smoothing.  
Red circles mark the isolated defect pixel.  
Bottom row: (d) pixel map of user activity with rogue nodes,  
(e) Laplacian response at $T=0.9$, and (f) neuron-model response.}
  \label{fig:real_images_with_filter_responses}
  \end{figure}

%
%
Next, is a rather specialised application of isolated pixel detection in image processing. Figure \ref{fig:real_images_with_filter_responses}(a) shows an X-ray image of a helicopter turbine blade with a porous region containing a tiny defect near the top right corner \cite{gonzalez2008digital}. This image has natural noise and texture making it a more difficult task than previous examples. The Laplacian method successfully detects the isolated pixel with a threshold of $0.9$. However, the neuron model required the application of a Gaussian $5\times5$ filter beforehand to avoid false positives, where pixels with significant local contrast were `incorrectly' flagged. (Note that Gaussian filtering is a common pre-processing step for imaging tasks, such as X-ray, MRI, or CT scans, typically using a $3\times3$ or $5\times5$ kernel to preserve fine details.)

Finally, a cybersecurity example can also be viewed as an image processing problem. Figure \ref{fig:real_images_with_filter_responses}(d) depicts a 2D pixel map of average user activity derived from spatially clustered communities, where neighbouring regions exhibit similar intensity values. Rogue users are visually represented as isolated pixels with anomalously high or low intensities — some more conspicuous than others. Figure \ref{fig:real_images_with_filter_responses}(e) shows the Laplacian method's response map with threshold of $T=0.9$, which detects only 2 isolated pixels. By manually lowering the threshold to $T=0.4$, all $10$ rogue users are detected. However, this threshold is highly sensitive to variations across images with different rogue user placements. In contrast, the proposed neuron model successfully identifies all rogue nodes without requiring any parameter tuning.

\section{Conclusion and future work}
\label{section:conclusion_and_future_work}
Most traditional image change detection methods are built on the foundation of derivatives, with the primary objective often being to determine whether a \emph{sufficient} change has occurred. Thresholding derivative outputs is a commonly used practice to establish decision boundaries; however, this shifts the challenge to the problem of selecting an appropriate threshold. In the work by \citet{nassir2021anomaly} a novel approach to global change detection was introduced by detecting anomalies using expectations; later represented as a non-linear neuron model and neural network \cite{mohammad2022net}. The present work modifies and extends the neuron model to detect localised anomalies, with the problem of detecting isolated pixels providing a concrete and focused application for understanding broader concepts in change detection. The neuron model is inspired by the human retina, where photoreceptors capture and count quanta of light, processing image intensities as a stream of discrete events. Rather than relying on first and second-order derivatives, change detection in a neurons receptive field is framed as an anomaly detection problem using lateral inhibition for specificity.

Experimental results demonstrate the effectiveness of the neuron model in detecting isolated pixels in binary images, and this success extends to artificial grayscale images with uniform intensity regions. While template matching performs well for binary images, it becomes impractical for grayscale images, and the Laplacian method requires manual threshold tuning, which may yield inconsistent results. The neuron model also demonstrates its potential in two further tasks of increased complexity. First, in detecting isolated pixels in X-ray imagery, smoothing the image before processing improved detection. This can be interpreted as providing additional constraints to the problem so as to detect the desired novelty in the image, and also follows a natural spatial filtering pre-processing step in image processing. Second, in the cybersecurity domain, the neuron model identified rogue nodes in network data, where these nodes manifest as isolated pixels within regions of similar values and noise.

Future research will enhance the model's capabilities by enabling it to operate across multiple scales; employing larger receptive fields to detect clusters of isolated pixels. Although the neuron computation is fast, optimisation efforts on the parallel processing will also be made to increase speed and efficiency. Additionally, expanding the model's capacity to detect more complex image features, such as lines and edges, will be explored to develop more robust and comprehensive feature detection models that do not reply on computing derivatives.


\bibliography{Bibliography.bib} 
\bibliographystyle{plainnat}

\end{document}